\documentclass[fonts]{icst}

\usepackage{moreverb}
\usepackage[breaklinks,colorlinks,bookmarksopen,bookmarksnumbered,linkcolor=ICSTblue,citecolor=blue,urlcolor=ICSTblue]{hyperref}
\usepackage{breakurl}
\usepackage{doi}
\usepackage{tabularx}
\usepackage{subcaption}
\usepackage{float}
\usepackage{graphicx,color,flushend}

\newcommand\BibTeX{{\rmfamily B\kern-.05em \textsc{i\kern-.025em b}\kern-.08emT\kern-.1667em\lower.7ex\hbox{E}\kern-.125emX}}

\def\volumeyear{2025}

\journalname{XXXXXX}
\articletype{Research Article/Editorial}
 \def\volumeyear{2025}
 \def\volumenumber{XX}
  \def\issuemonth{XX-XX}
  \def\issuenumber{X}
  \def\articleid{XX}{}
  \copyrightnote{This is an open access article distributed under the terms of the Creative Commons Attribution license (\url{http://creativecommons.org/licenses/by/3.0/}), which permits unlimited use, distribution and reproduction in any medium so long as the original work is properly cited.}
  \def\articledoi{10.4108/XX.X.X.XX}
  \received{August 2025}
  \accepted{XXXX}
  \published{XXXX}
\begin{document}

\runningheads{}{}

\title{SmartFlow: Reinforcement Learning and Agentic AI for Bike-Sharing Optimisation} 

\author{Aditya Sreevatsa K, Arun Kumar Raveendran, Jesrael K Mani, Prakash G Shigli, Rajkumar Rangadore, Narayana Darapaneni, Anwesh Reddy Paduri\fnoteref{1}}

\address{Department of Data Science, Machine Learning \& Artificial Intelligence, PES University, Bengaluru, Karnataka - 560100, India}

\abstract{
SmartFlow is a multi-layered framework that integrates \textit{Reinforcement Learning and Agentic AI} to address the dynamic rebalancing problem in urban bike-sharing services. Its architecture separates strategic, tactical, and communication functions for clarity and scalability. At the strategic level, a \textbf{Deep Q-Network (DQN) agent}, trained in a high-fidelity simulation of New York’s Citi Bike network, learns robust rebalancing policies by modelling the challenge as a Markov Decision Process. These high-level strategies feed into a deterministic tactical module that optimises multi-leg journeys and schedules just-in-time dispatches to minimise fleet travel. Evaluation across multiple seeded runs demonstrates SmartFlow’s high efficacy, \textbf{reducing network imbalance by over 95\%} while requiring minimal travel distance and achieving strong truck utilisation. A communication layer, powered by a grounded Agentic AI with a Large Language Model (LLM), translates logistical plans into clear, actionable instructions for operational staff, ensuring interpretability and execution readiness. This integration bridges machine intelligence with human operations, offering a scalable solution that reduces idle time, improves bike availability, and lowers operational costs. SmartFlow provides a blueprint for interpretable, AI-driven logistics in complex urban mobility networks.
}


\fnotetext[1]{Corresponding author.  Email: \email{anwesh@greatlearning.in}}
\maketitle

\section{Introduction}
\label{sec:introduction}

Bike-Sharing Systems (BSS) have emerged as a cornerstone of modern urban mobility, offering a convenient and sustainable solution to the perennial challenge of last-mile transportation. By providing fleets of bicycles for short-term hire, these systems contribute significantly to reducing traffic congestion, promoting healthier lifestyles, and supporting integrated public transport networks. As cities worldwide intensify their focus on sustainability, the effective management of these shared resources has become a matter of critical importance.

Yet, the promise of these systems is frequently undermined by a persistent operational hurdle: the dynamic imbalance of bicycle availability. Caused by tidal flows of commuters and fluctuating demand patterns, this imbalance results in stations being either completely empty at peak times or frustratingly full, which diminishes service quality and inflates operational expenditure. Conventional rebalancing methods, which rely on moving bicycles with fleets of service vehicles, are often static or manually scheduled. They fail to adapt to the fluid reality of urban dynamics and incur significant logistical costs related to fuel, vehicle maintenance, and personnel, while offering only a partial solution to the problem.

This challenge coincides with a transformative era in artificial intelligence, marking the arrival of a fifth generation of BSS that leverages intelligent, data-driven technologies. With advancements in Reinforcement Learning (RL), there is a profound opportunity to move beyond reactive methods and instead develop an agent that can learn an optimal, forward-looking rebalancing policy through direct interaction with its environment. Furthermore, the rise of Agentic AI presents a novel way to bridge the gap between complex machine-generated strategies and their real-world human execution.

In response to this opportunity, this paper introduces \textbf{SmartFlow}, a hybrid framework that synergises deep reinforcement learning with agentic AI to deliver an intelligent and proactive rebalancing solution. At its core, a Deep Q-Network (DQN) agent is trained within a high-fidelity simulation environment to learn a strategic policy for station-level bike redistribution. This separation of concerns allows the RL agent to focus on the high-level strategic goal of network balance, while a deterministic tactical planning module handles the grounded complexities of optimising multi-leg journeys for service vehicles. Finally, an innovative agentic AI layer autonomously translates these optimised plans into clear, human-readable dispatch instructions, reducing the potential for ambiguity and ensuring the agent's strategy is executed with precision.

The primary contributions of this work are threefold: first, we model the dynamic bike rebalancing problem within a bespoke reinforcement learning environment; second, we develop and train a DQN agent capable of recommending optimal redistribution actions that significantly reduce network imbalance; and third, we demonstrate the framework's ability to improve operational efficiency and provide a scalable, transparent solution for managing complex urban mobility systems.

\section{Related Work}
\label{sec:related_work}

The operational challenge of maintaining a balanced distribution of bicycles in a Bike-Sharing System (BSS) is a well-established area of research, broadly categorised into static and dynamic rebalancing problems. Early works predominantly focused on the Static Bike Rebalancing Problem (SBRP), where vehicle routes are optimised during off-peak hours based on historical demand. These studies often formulated the task as a capacitated vehicle routing problem, employing methods such as Mixed Integer Linear Programming (MILP) to find optimal solutions \cite{benchimol2011balancing, chemla2013bike}. While foundational, static approaches lack the flexibility to respond to real-time demand fluctuations, a critical requirement for modern, large-scale systems.

Consequently, research has shifted towards the Dynamic Bike Rebalancing Problem (DBRP), which seeks to make adaptive decisions based on the current state of the network. A wide array of methodologies has been applied to this more complex problem. Mathematical optimisation continues to be relevant, with stochastic and robust optimisation models designed to handle demand uncertainty, though they often face challenges with computational scalability \cite{raviv2013static, bruck2024robust}. To overcome these scaling issues, many studies have proposed heuristic and simulation-based strategies. These methods, including agent-based simulation and curvature-based algorithms, offer practical and scalable solutions by prioritising near-optimal decisions for real-time implementation \cite{ban2019curvature, ghosh2017dynamic}. Other data-driven approaches have leveraged graph theory and community detection to simplify the problem by clustering stations, or used computer vision techniques on mobility heatmaps to enhance demand forecasting \cite{Li2021Dynamic, yang2020towards}.

More recently, Reinforcement Learning (RL) has emerged as a particularly promising paradigm for the DBRP. RL is inherently suited for sequential decision-making in complex, uncertain environments, as it allows an agent to learn an effective policy directly from interaction without needing an explicit model of the system \cite{sutton2018reinforcement}. Several works have demonstrated the superiority of deep RL methods over traditional heuristics. For instance, recent approaches have integrated Graph Neural Networks with Deep Q-Networks (DQN) in frameworks like DeepBike \cite{Zheng2023DeepBike}, or have explored Multi-Agent Reinforcement Learning (MARL) to coordinate fleets of vehicles, showing significant improvements in service levels and reductions in lost user demand \cite{Zhou2023STCBR, liang2024dual}.

The trend towards hybrid frameworks, which combine the strengths of different techniques, underscores the complexity of the rebalancing task \cite{li2023mlpga, wang2021fleet}. Our work contributes to this line of research by proposing SmartFlow, a novel hybrid system that addresses a crucial, yet often overlooked, aspect of real-world deployment: the communication gap between an AI's optimised strategy and the human operators who must execute it. While other systems focus solely on deriving an optimal policy, SmartFlow integrates its RL agent with a modern Agentic AI layer. This layer leverages a Large Language Model to autonomously translate the computed rebalancing plan into clear, human-readable instructions, a concept inspired by recent work in language-action models \cite{yao2022react}. This synergistic approach aims not only to optimise logistics but also to ensure the solution is transparent, auditable, and seamlessly integrated into human-in-the-loop operations.

\subsection{Hybrid Frameworks in Bike Rebalancing}

Recognising that no single method perfectly solves the rebalancing problem, recent research has focused on hybrid frameworks that combine the strengths of different AI and operational research paradigms. These systems often pair a predictive or learning-based component with a deterministic optimisation algorithm.

One common approach is to couple machine learning-based demand prediction with a subsequent optimisation step. Lin et al. (2018) \cite{lin2018graph}, for example, first use a Graph Convolutional Network (GCNN) to predict station-level demand, and then feed these predictions into a capacitated location-routing model to plan vehicle movements. Similarly, Li et al. (2023) \cite{li2023mlpga} use a Multilayer Perceptron (MLP) for forecasting, but then employ a Genetic Algorithm (GA) to find an optimal mix of truck-based moves and user-based incentives, effectively optimising over two different rebalancing strategies simultaneously.

Another powerful hybrid paradigm involves integrating heuristics with high-fidelity simulations. Ban et al. (2019) \cite{ban2019curvature} developed a "curvature map" heuristic, where bike surpluses and deficits are treated as a 3D terrain. A greedy algorithm then generates routes that move bikes from convex "hills" to concave "craters". This lightweight heuristic is embedded within an agent-based simulator that provides an online feedback loop, allowing subsequent routes to be adjusted based on the freshest system state.

The trend towards hybrid frameworks, which combine the strengths of different techniques, underscores the complexity of the rebalancing task. For instance, researchers have paired predictive models like Graph Convolutional Networks with traditional routing algorithms \cite{lin2018graph}, or have developed dual-policy RL systems that decouple strategic inventory control from tactical routing \cite{liang2024dual}. Our work contributes to this line of research by proposing SmartFlow, a novel hybrid system that addresses a crucial, yet often overlooked, aspect of real-world deployment: the communication gap between an AI's optimised strategy and the human operators who must execute it. While other systems focus solely on deriving an optimal policy, SmartFlow integrates its RL agent with a modern Agentic AI layer. This layer leverages a Large Language Model to autonomously translate the computed rebalancing plan into clear, human-readable instructions, a concept inspired by recent work in language-action models \cite{yao2022react}. This synergistic approach aims not only to optimise logistics but also to ensure the solution is transparent, auditable, and seamlessly integrated into human-in-the-loop operations.

\section{Materials and Methods}
\label{sec:materials_methods}

The SmartFlow framework is architected as a multi-layered system designed to translate high-level strategic learning into tactical, real-world action. It synergises data engineering, deep reinforcement learning, and agentic AI to create a comprehensive pipeline, from data preparation to autonomous operational execution. The complete implementation, encompassing preprocessing pipelines, model training procedures, and visualisation components, is accessible in the SmartFlow repo \footnote{SmartFlow Repo: https://github.com/AdityaSreevatsaK/SmartFlow}~\cite{smartflow_repo}.

\subsection{Theoretical Framework}

The design of SmartFlow is grounded in two core areas of artificial intelligence: value-based reinforcement learning and agentic language models.

\paragraph{Value-Based Reinforcement Learning.}
Value-based RL methods are designed to solve the problem defined by the MDP by learning an optimal \textbf{action-value function}, $Q^*(s, a)$. This function estimates the maximum expected future reward achievable from taking action $a$ in state $s$ and continuing optimally thereafter \cite{sutton2018reinforcement}. The function adheres to the Bellman optimality equation:
\begin{equation}
    Q^*(s,a) = \mathcal{R}(s,a) + \gamma \sum_{s'} \mathcal{P}(s'|s,a) \max_{a'} Q^*(s',a')
\end{equation}
For problems with a large or continuous state space, such as bike rebalancing, it is intractable to represent this function as a table. The \textbf{Deep Q-Network (DQN)} algorithm overcomes this by using a deep neural network with weights $\theta$ as a powerful function approximator to estimate the action-value function, $Q(s, a; \theta)$ \cite{mnih2015human}. To ensure stable training, DQN introduces two crucial innovations. First, \textbf{Experience Replay} stores past transitions in a replay buffer and samples mini-batches from it to train the network. This breaks the temporal correlations in sequential observations, satisfying the i.i.d. assumption required for stable deep learning. Second, a separate \textbf{Target Network}, with weights that are only periodically updated, is used to generate the Bellman targets. This decouples the target value from the online network's weights, preventing the oscillatory and divergent learning patterns that can otherwise occur.

\paragraph{Agentic AI and Grounded Reasoning.}
An "agentic" model is a system capable of reasoning and acting to achieve goals. While the RL agent learns an optimal numeric policy, the agentic AI layer handles the final, crucial step of communicating the system's plan to human operators \cite{yao2022react}. A primary theoretical challenge in using Large Language Models for such tasks is ensuring factual consistency and mitigating the risk of "hallucination." SmartFlow addresses this by employing \textbf{grounded prompt engineering}, where the LLM is explicitly constrained through its prompt to reason \textit{only} over the provided data from the planning module \cite{huang2022language}. This ensures the agent's natural language output is a reliable and verifiable translation of the optimised plan, rather than an unverified invention.

\subsection{System Architecture}

\begin{figure}[H]
  \centering
  \includegraphics[width=\columnwidth]{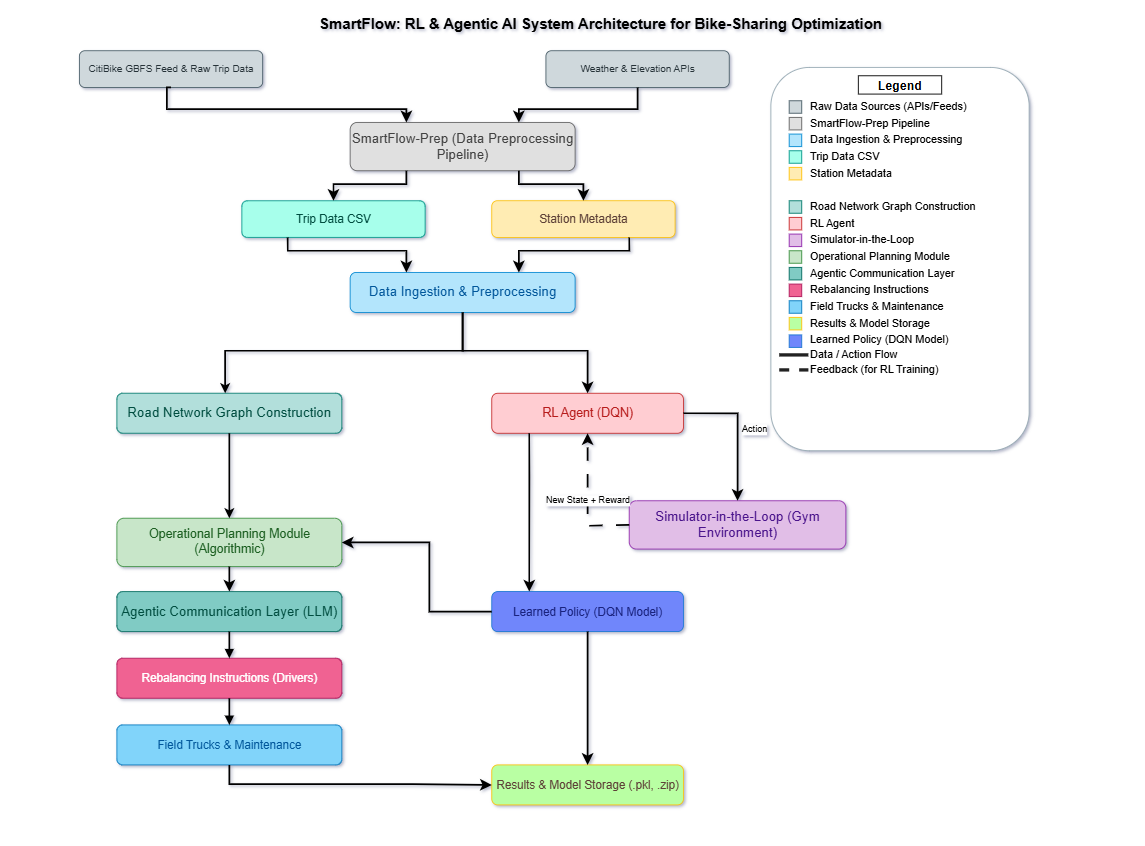}
  \caption{System architecture of the SmartFlow framework. A strategic RL agent guides a tactical planning module, whose output is translated into human-readable instructions by an agentic AI layer.}
  \label{fig:architecture}
\end{figure}

The SmartFlow framework is architected with a deliberate separation of concerns, decomposing the complex rebalancing task into distinct strategic, tactical, and communication layers. This multi-layered design, depicted in Figure~\ref{fig:architecture}, allows each component to focus on a specific part of the problem, synergising deep reinforcement learning and agentic AI to create a comprehensive pipeline from data preparation to autonomous operational execution.

\paragraph{Strategic Core: The Reinforcement Learning Agent.}
The high-level intelligence of the framework is a Deep Q-Network (DQN) agent that functions as the system's \textbf{strategist}. This agent's primary responsibility is to learn a long-term, optimal rebalancing policy. To achieve this, it operates within a high-fidelity \textbf{Simulator-in-the-Loop}, which serves as a digital twin of the urban environment. Within this safe and scalable simulator, the agent can explore millions of state-action possibilities through trial-and-error, learning from the consequences of its decisions without incurring real-world costs or service disruptions \cite{ban2019curvature}. The output of this layer is not a rigid set of commands, but a high-level strategic policy that maps any given network state to the most advantageous bike transfer required to pre-emptively balance the system.

\paragraph{Tactical Execution: The Operational Planning Module.}
The abstract strategy developed by the RL agent is then passed to the \textbf{tactician}: a deterministic Operational Planning Module. This module acts as the crucial bridge between the learned policy and concrete logistics. It ingests the series of strategic transfers recommended by the agent and executes a two-step process. First, a multi-leg journey optimisation algorithm chains individual transfers into efficient, continuous routes for the vehicle fleet, maximising the utility of each dispatch. Second, a just-in-time scheduling algorithm assigns a proactive dispatch time to each leg of the journey, ensuring that bikes are moved just before they are needed.

\paragraph{Communication Interface: The Agentic AI Layer.}
The final, and most novel, layer is the \textbf{Agentic Communication Layer}, which is responsible for operationalising the tactical plan. It receives the optimised journey schedule and uses a Large Language Model (LLM) to translate the structured, machine-readable data into unambiguous, human-readable dispatch instructions. This ensures the complex, AI-generated strategy is rendered in a format that is transparent, auditable, and immediately actionable by human crews in the field, a critical step for ensuring the framework's practical utility and adoption \cite{yao2022react}.

\subsection{Data Acquisition and The SmartFlow-Prep Pipeline}

The framework was trained and evaluated on a public dataset from New York's Citi Bike programme, containing trip logs from 2015 to 2017. To process this raw data, a dedicated pipeline, \textbf{SmartFlow-Prep}\footnote{Repo link: https://github.com/AdityaSreevatsaK/SmartFlow-Prep}, was engineered to systematically transform heterogeneous data sources into a clean, feature-rich, and analysis-ready format \cite{SmartFlow-Prep}. This pipeline executes a sequence of modular scripts to ensure data integrity and consistency.

The process begins by sourcing foundational data. Station metadata, including names, capacities, and geographic coordinates, is fetched from the official CitiBike General Bikeshare Feed Specification (GBFS) API. Concurrently, historical daily weather data for New York City, including temperature and precipitation, is collected from the Visual Crossing Weather API for the entire three-year period.

Next, the pipeline performs data enrichment. A temporal feature engineering script parses all trip timestamps to extract the date, hour, weekday, and a weekend/holiday flag. This is followed by spatial enrichment, where an open-elevation API is queried to append the elevation of each station, providing a proxy for route difficulty. Finally, a systematic data cleansing script is run. It removes records with missing coordinates or invalid trip durations and applies logical filters to ensure data integrity. The fully cleaned and enriched trip data is then merged with the weather and station metadata to produce the final, holistic dataset used for model training.

\subsection{Reinforcement Learning Environment}

A bespoke simulation environment, engineered using the Gymnasium library, serves as a high-fidelity digital twin of the bike-sharing network. Its primary purpose is to provide a safe, scalable, and computationally efficient arena for the RL agent to learn through millions of trial-and-error interactions—a process that would be prohibitively expensive and disruptive in the real world. The environment encapsulates the system's state, permissible actions, and dynamic transitions, providing the essential feedback loop for policy learning.

\paragraph{State and Action Spaces.}
The agent's interface with the environment is formally defined by its observation and action spaces. The \textbf{observation space} is a vector containing the real-time bike count for every station, plus an additional value representing the current hour of the day (0-23). The inclusion of the temporal feature is critical, as it allows the agent to learn time-dependent policies that anticipate cyclical demand patterns like morning and evening commutes. The \textbf{action space} is discrete, comprising every possible transfer of a single bike between any two distinct stations in the network, forming the set of all potential rebalancing moves.

\paragraph{Environment Dynamics and State Transitions.}
The core logic of the simulation is managed by the \texttt{step()} method, which executes a single time step. When the agent selects an action, the environment first validates its feasibility; a transfer is only permitted if the source station has a surplus of bikes and the target station has available capacity. If the action is valid, the bike counts are updated. Subsequently, the environment simulates one hour of public usage by adjusting the inventories at all stations based on the pre-processed historical demand data. After this public demand is applied, station capacity constraints are enforced (i.e., inventories are clipped to their minimum and maximum). Finally, a reward is calculated based on the outcome of the agent's action, the simulation clock is advanced by one hour, and the new state and reward are returned to the agent.

\paragraph{Episodic Reset for Robust Learning.}
At the beginning of each training episode, the \texttt{reset()} method is called to prepare the environment for a new run. This function returns the environment to a fresh, randomised initial state by generating new starting inventory levels for each station. This randomisation is a crucial part of the training regime. By exposing the agent to a wide variety of starting conditions, it prevents the agent from overfitting to a specific initial scenario and forces it to learn a more robust and generalisable policy that is effective across many different network states.

\subsection{Problem Formulation}
The dynamic rebalancing problem is formally modelled as a Markov Decision Process (MDP), defined by the tuple $\mathcal{M} = (S, A, P, R, \gamma)$ \cite{sutton2018reinforcement, puterman1994markov}. At each discrete hourly time step $t$, the agent observes the system state $s_t$, selects an action $a_t$, receives a reward $r_{t+1}$, and transitions to a new state $s_{t+1}$. The components are defined as:

\paragraph{State ($s_t \in S$):} The state is a vector $s_t = (I_t, T_t)$, where $I_t \in \mathbb{Z}_+^N$ is a vector representing the number of available bikes at each of the $N$ stations, and $T_t \in \{0, ..., 23\}$ is the current hour of the day.

\paragraph{Action ($a_t \in A$):} The action space is discrete, where a single action $a_t = (i, j)$ represents the transfer of one bike from a source station $i$ to a destination station $j$.

\paragraph{Reward ($R(s_t, a_t)$):} A shaped reward function guides the agent's learning. A positive reward is given for moving a bike to a station in need, scaled by the magnitude of that need. Significant negative penalties are applied for infeasible actions, with smaller penalties for inefficient but feasible moves.

The agent's objective is to learn an optimal policy $\pi^*$ that maximises the expected cumulative discounted reward:
\begin{equation}
   \pi^* = \arg\max_\pi \mathbb{E}_\pi \left[ \sum_{t=0}^T \gamma^t R(s_t, a_t) \right]
\end{equation}

\subsection{Model Implementation}

The SmartFlow framework was implemented in Python. The historical trip data was sourced from New York City's official Citi Bike programme, focusing on the top 30 busiest stations for the simulation experiments.

\paragraph{Reinforcement Learning Agent.}
The DQN agent's architecture consists of a Multi-Layer Perceptron (MLP) with two hidden layers of 128 neurons each, providing sufficient capacity to learn the complex relationships between the state variables. Key hyperparameters were carefully selected to ensure stable convergence: a learning rate of \(1 \times 10^{-4}\) was used for gradual policy updates, and a large experience replay buffer of 50,000 transitions was implemented to break temporal correlations and improve data efficiency. The model begins learning after an initial 1,000 steps of exploration and updates its network every four steps. To stabilise the learning targets, a separate target network was used, with its weights updated periodically to match the main network, a standard practice for robust DQN training \cite{mnih2015human}.

\subsubsection{Agentic AI Communication Layer}
The final and most novel component of the framework is the agentic AI layer, which serves as the crucial interface between the system’s optimised plan and the human operators responsible for its execution. Its purpose is to translate the structured, machine-readable journey plan into a clear, natural-language dispatch report, thereby enhancing interpretability and usability \cite{yao2022react}.

This module is powered by the \texttt{google/gemma-2b-it} Large Language Model (LLM), accessed via a text-generation pipeline from the \texttt{transformers} library. To ensure the output is factually accurate and reliable, we employ a strategy of \textbf{grounded prompt engineering}. A detailed, multi-part prompt is provided to the LLM at inference time, which defines its behaviour:
\begin{itemize}
    \item \textbf{Persona and Task:} The prompt instructs the LLM to adopt the persona of "SmartFlow, an autonomous Logistics Analyst" and assigns it the explicit task of generating a manager's briefing and per-truck dispatch tickets.
    \item \textbf{Data Grounding:} Critically, the prompt includes the complete, JSON-formatted journey plan and contains a non-negotiable rule that the LLM must generate its report using \textit{only} the provided data. This constraint is fundamental to mitigating the risk of model hallucination and ensuring the final report is a verifiable representation of the optimised plan \cite{huang2022language}.
    \item \textbf{Format Specification:} The prompt specifies the exact Markdown structure required for the output, ensuring the final report is consistently formatted and easy to parse.
\end{itemize}
To guarantee operational robustness, the entire LLM call is wrapped in a fallback mechanism. In the event of an API or model failure, the system defaults to a deterministic Python formatter that produces a simpler, but still structured and usable, dispatch ticket. This dual approach provides both the sophisticated, human-like reporting of an LLM and the resilience of a traditional system.

\subsection{Operational Planning and Scheduling}
Once the RL agent has produced a strategic plan of required bike transfers, the framework transitions to tactical execution via the Operational Planning Module. This module uses two deterministic algorithms to create an efficient plan for the vehicle fleet.

First, a \textbf{multi-leg journey optimisation} algorithm converts the simple list of transfers into efficient, chained journeys. It iteratively identifies the station with the greatest surplus, 'loads' a virtual truck with all available bikes, and plans a sequential route to the nearest stations with deficits. This process repeats until the truck's inventory is depleted, a method designed to minimise vehicle mileage and the total number of dispatches.

Second, a \textbf{just-in-time dispatch scheduling} algorithm assigns a proactive schedule to each journey. It prioritises tasks based on the urgency determined by the RL agent (i.e., the hour of identified need). The algorithm then works backwards from this time, subtracting estimated travel durations to calculate an ideal dispatch time for each leg of the journey, ensuring that resources are allocated efficiently and bikes arrive just before they are needed.

\subsection{Generalisation and Cross-City Deployment}
A critical consideration for any BSS framework is its ability to generalise across different urban environments without requiring complete retraining. The SmartFlow architecture is designed with this in mind. The use of a deep neural network allows the agent to learn abstract representations of network states rather than memorising specific inventory levels for one city. This learned policy could be used as a powerful starting point for a new city deployment via \textbf{transfer learning} \cite{taylor2009transfer, lazaric2012transfer}. By fine-tuning the pre-trained model on a small amount of data from a new city, the agent could adapt much more quickly than one learning from scratch, significantly reducing the cost and time of new deployments.

\subsection{Practical Feasibility and Deployment Considerations}
While this study is grounded in a high-fidelity simulation, the SmartFlow framework has been designed with practical deployment in mind. The feasibility of a real-world implementation is supported by several factors. Firstly, the \textbf{data availability} is high; the required inputs, such as historical trip logs and live station status, are commonly accessible from BSS operators through public datasets or APIs.

Secondly, the \textbf{computational resources} required are manageable. While training a deep reinforcement learning model is computationally intensive, it is an offline process that can be performed periodically on cloud-based GPU instances. The trained DQN model is highly efficient for inference, allowing for real-time action selection on standard hardware.

Finally, the framework is designed for seamless \textbf{operational integration}. The agentic AI layer's output is a structured, natural-language report that can be directly consumed by existing fleet management software or human dispatchers. This allows for a flexible human-in-the-loop approach, where the system provides intelligent recommendations that can be reviewed and actioned by operational staff, thereby lowering the barrier to adoption. The primary challenge remains the simulation-to-reality gap, which can be addressed in a live deployment through continuous online learning and adaptation.

\subsection{End-to-End Simulation Workflow}
The SmartFlow framework is executed via a central orchestration script that manages the entire pipeline in a sequential, four-phase process.

\paragraph{Phase 1: Initialisation.} The workflow begins by loading the pre-processed trip and station data from the \textit{SmartFlow-Prep}~\cite{SmartFlow-Prep} pipeline. It then constructs the geospatial road network graph using OSMnx and initialises the Gymnasium simulation environment to a randomised starting state of bike inventories for the specified target date.

\paragraph{Phase 2: Strategic Planning via RL.} The main simulation loop commences, advancing in discrete hourly timesteps. At each step, the pre-trained DQN agent observes the current network state (station inventories and time of day) and selects a strategic transfer action $(i, j)$ that maximises its learned Q-value function. The environment executes the action, simulates public demand for that hour, and returns a new state and reward. This loop continues until the end of the 24-hour simulation period, resulting in a complete strategic plan composed of all the actions the agent took.

\paragraph{Phase 3: Tactical Plan Generation.} After the simulation is complete, the high-level strategic plan is passed to the Operational Planning Module. This module first applies the multi-leg journey optimisation algorithm to chain the individual transfers into efficient, continuous routes for a fleet of virtual trucks. Subsequently, the just-in-time scheduling algorithm is applied to assign a proactive dispatch time to each leg of every journey, creating a final, deconflicted tactical plan.

\paragraph{Phase 4: Reporting and Visualisation.} In the final phase, the tactical plan is passed to the Agentic AI layer, which uses the Gemma LLM to generate the final human-readable dispatch report. Concurrently, the same plan is used by the visualisation module to render the comprehensive and interactive Folium map, plotting the station states and animating the optimised truck routes.

\paragraph{Multi-Run Execution for Statistical Robustness.} The four-phase process described above constitutes a single, end-to-end experimental run. To ensure the statistical validity of the final results, this entire workflow is invoked by a master script that iterates through a predefined list of random seeds. For each seed, a complete and independent simulation is conducted, generating a distinct set of results and providing the foundation for the aggregated statistical analysis.
\section{Results and Analysis}
\label{sec:results}

To evaluate the SmartFlow framework, we conducted a series of experiments based on historical data from New York's Citi Bike network. The simulation was focused on the top 30 busiest stations on a representative day (1st July 2016). To ensure statistical validity, the entire pipeline was executed across \textbf{three independent, seeded runs}, and the following results are presented as both individual run details and aggregated statistics.

\subsection{Agent Training and Convergence}
\begin{figure}[H]
    \centering
    \includegraphics[width=\linewidth]{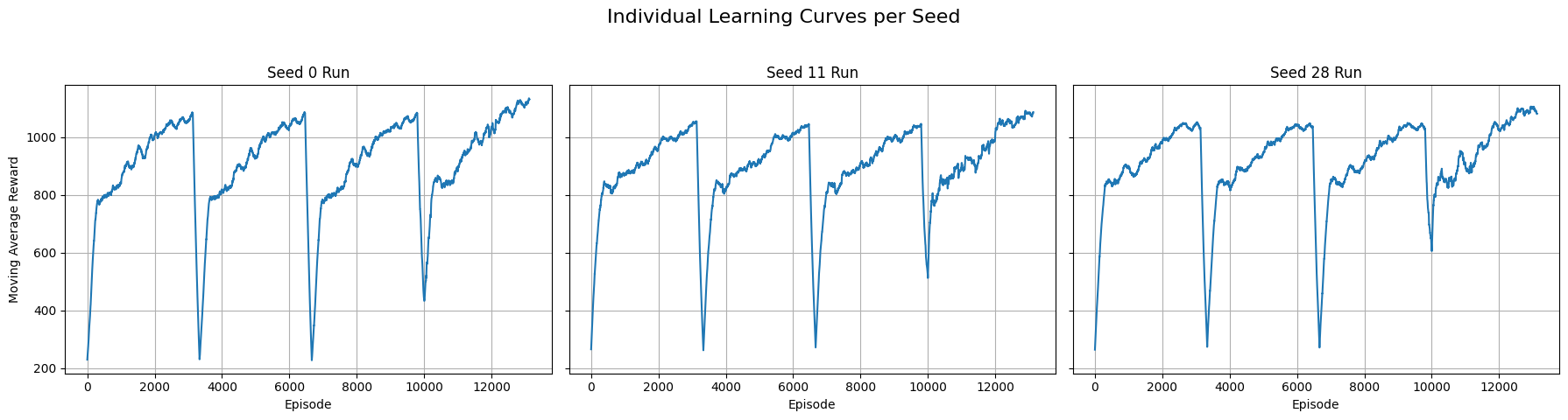}
    \caption{Individual learning curves for each of the three seeded runs, showing the moving average reward over time.}
    \label{fig:ind_learning_curves}
\end{figure}

The Deep Q-Network agent was trained for one million timesteps in each of the three runs. The agent's learning progress and stability are visualised by first examining the reward curves from each independent run (Fig.~\ref{fig:ind_learning_curves}). While each run exhibits the natural variance and cyclical dips characteristic of reinforcement learning's exploration process, all three show a clear and positive learning trajectory.

\begin{figure}[H]
 \centering
 \includegraphics[width=0.8\columnwidth]{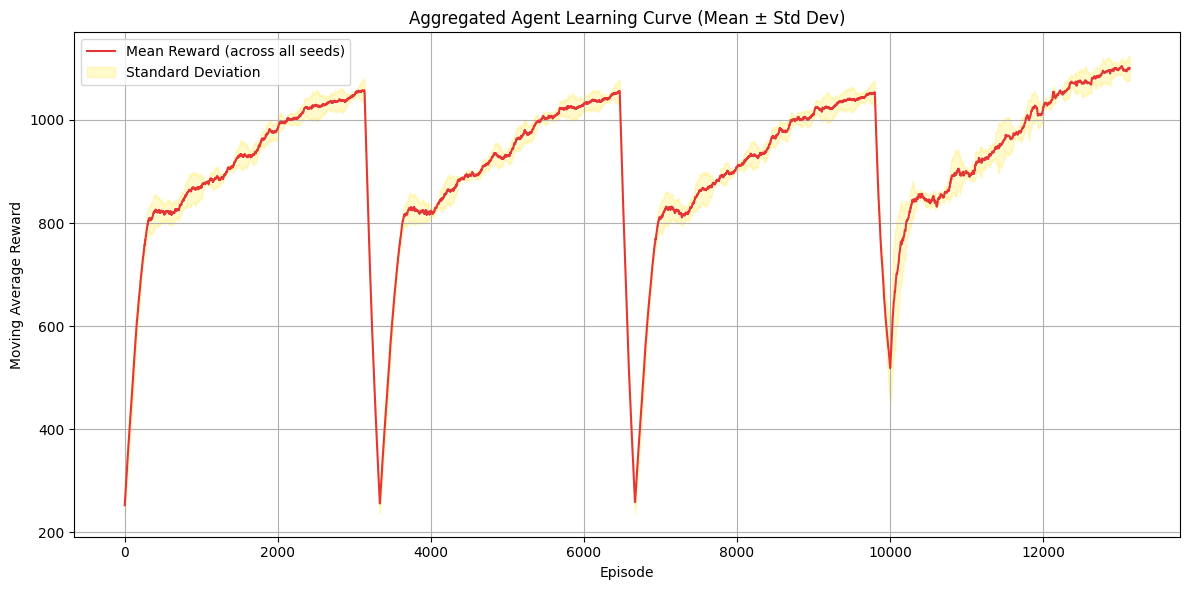}
 \caption{The aggregated learning curve across three runs, showing the mean moving average reward and standard deviation.}
 \label{fig:agg_learning_curve}
\end{figure}

To provide a more holistic view, these results are consolidated into a single aggregated plot (Fig.~\ref{fig:agg_learning_curve}). The mean reward (solid line) and standard deviation (shaded area) confirm that the agent consistently learned a robust policy, rather than succeeding due to a single "lucky" run. The aggregated training statistics are summarised in Table~\ref{tab:agg_results}, showing a final policy loss that indicates successful convergence across all runs.

\subsection{Operational Performance and Qualitative Analysis}
The framework's final performance, aggregated across all runs, is summarised in Table~\ref{tab:agg_results}. The agent's strategic policy proved to be highly effective, achieving a remarkable average \textbf{imbalance reduction of 95.47\%}. This indicates that the system successfully brought the network to a state of near-perfect equilibrium, directly addressing the core problem of station unavailability for end-users.

Crucially, this was accomplished with exceptional operational efficiency. The entire rebalancing operation required an average of only \textbf{37.21 km} of total fleet travel, demonstrating that the generated routes were both direct and effective. The intelligence of the tactical planning module is further confirmed by the high average \textbf{Truck Utilisation Rate of 66.15\%}. This result shows that for two-thirds of all dispatches, the system successfully chained multiple tasks into efficient, multi-leg journeys, where a single truck could service several stations in one trip. This capability is key to minimising the number of vehicles required and lowering operational costs.

\begin{table}[H]
\centering
\caption{Aggregated Performance Metrics Across Three Independent Runs.}
\label{tab:agg_results}
\begin{tabular}{ll}
\toprule
\textbf{Metric} & \textbf{Value (Mean \(\pm\) Std Dev)} \\
\midrule
\multicolumn{2}{l}{\textit{System-Level Performance}} \\
Imbalance Reduction & 95.47 \(\pm\) 2.73 \% \\
\midrule
\multicolumn{2}{l}{\textit{Operational Efficiency}} \\
Total Fleet Distance & 37.21 \(\pm\) 5.81 km \\
Truck Utilisation Rate & 66.15 \(\pm\) 25.50 \% \\
\midrule
\multicolumn{2}{l}{\textit{Training Convergence}} \\
Final Policy Loss & 3.4918 \(\pm\) 1.6791 \\
\bottomrule
\end{tabular}
\end{table}

\begin{figure}[H]
    \centering
    \includegraphics[width=\linewidth]{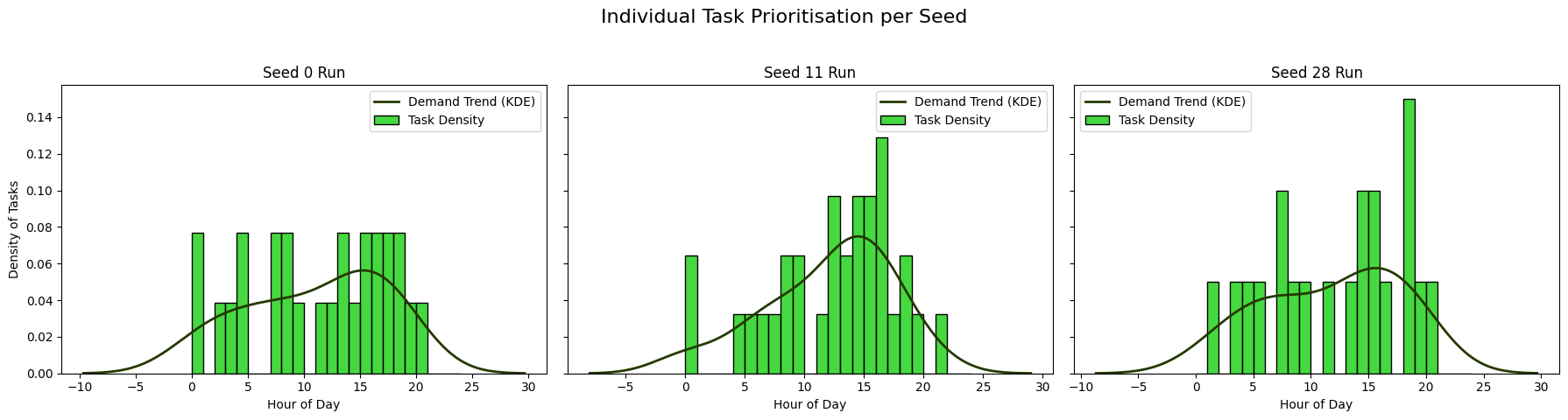}
    \caption{Individual task prioritisation plots for each of the three seeded runs.}
    \label{fig:ind_task_prio}
\end{figure}

Beyond these metrics, the agent consistently demonstrated sophisticated, proactive planning. Figure~\ref{fig:ind_task_prio} analyses the temporal dimension of each run, showing a consistent pattern of proactive decision-making. In all runs, the agent identifies the need for intervention during off-peak hours, particularly in the early morning and afternoon.

This proactive behaviour is first evident when examining the results from each seeded run, as shown in Figure~\ref{fig:ind_task_prio}. While the exact distribution of tasks varies slightly between seeds, a consistent pattern emerges: in all runs, the agent identifies the need for intervention during off-peak hours, particularly in the early morning and afternoon. This consistency across independent experiments demonstrates that the proactive strategy is a robust and replicable feature of the learned policy, not an incidental outcome.

\begin{figure}[H]
 \centering
 \includegraphics[width=0.8\columnwidth]{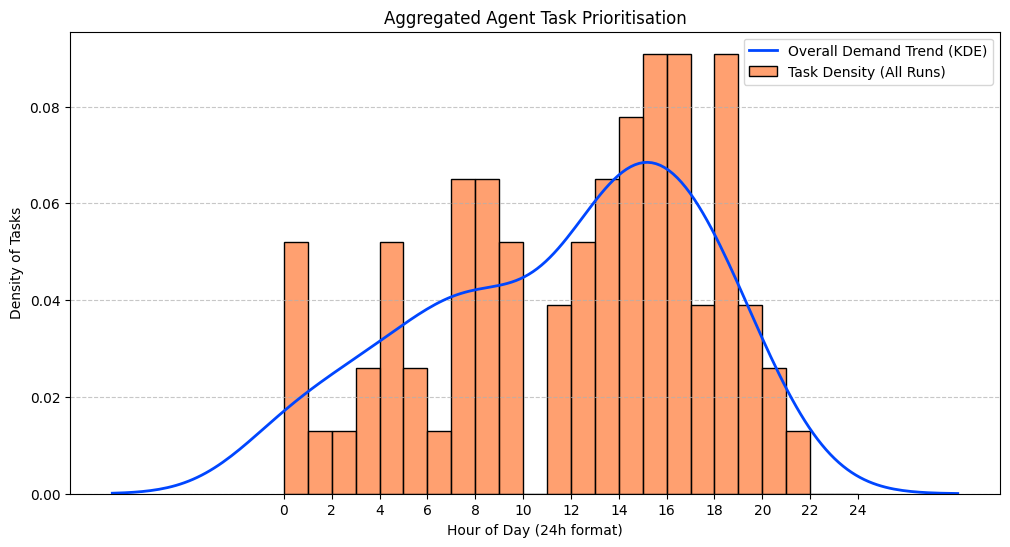}
 \caption{Aggregated density plot of the hours at which the agent identified the need for a transfer, showing a consistent pattern of proactive planning.}
 \label{fig:agg_task_prio}
\end{figure}

The aggregated data in Figure~\ref{fig:agg_task_prio} confirms and clarifies this sophisticated temporal strategy. The combined "Task Density" shows two clear peaks of activity. The first, in the early morning (e.g., 2-4 AM), represents the agent learning to pre-position bicycles during the network's quietest hours in anticipation of the morning rush several hours later. The second, larger peak in the afternoon (e.g., 2 PM) shows the agent proactively rebalancing the network ahead of the evening commute. This forward-looking behaviour contrasts sharply with a naive, reactive strategy, confirming that the agent has successfully learned to anticipate and mitigate future demand imbalances.

\begin{figure}[H]
    \centering
    \includegraphics[width=\linewidth]{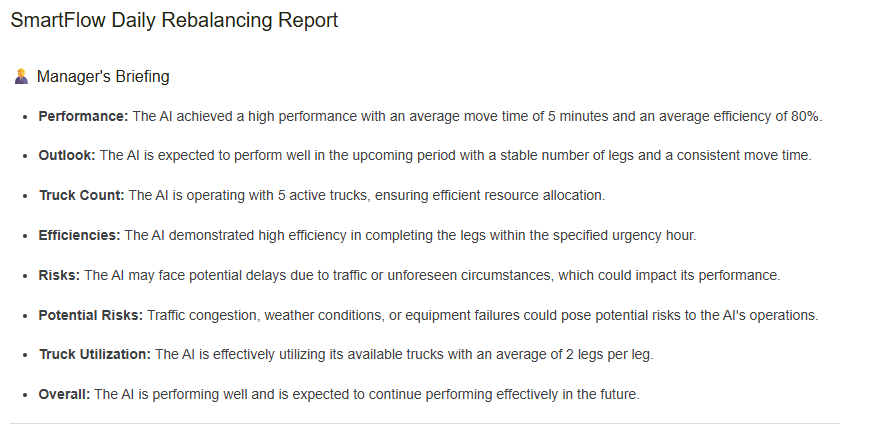}
    \caption{A representative output from the Agentic AI layer, showing the Manager's Briefing and a sample of the detailed, per-truck Dispatch Tickets.}
    \label{fig:agentic_report}
\end{figure}

\begin{figure}[H]
    \centering
    \includegraphics[width=\linewidth]{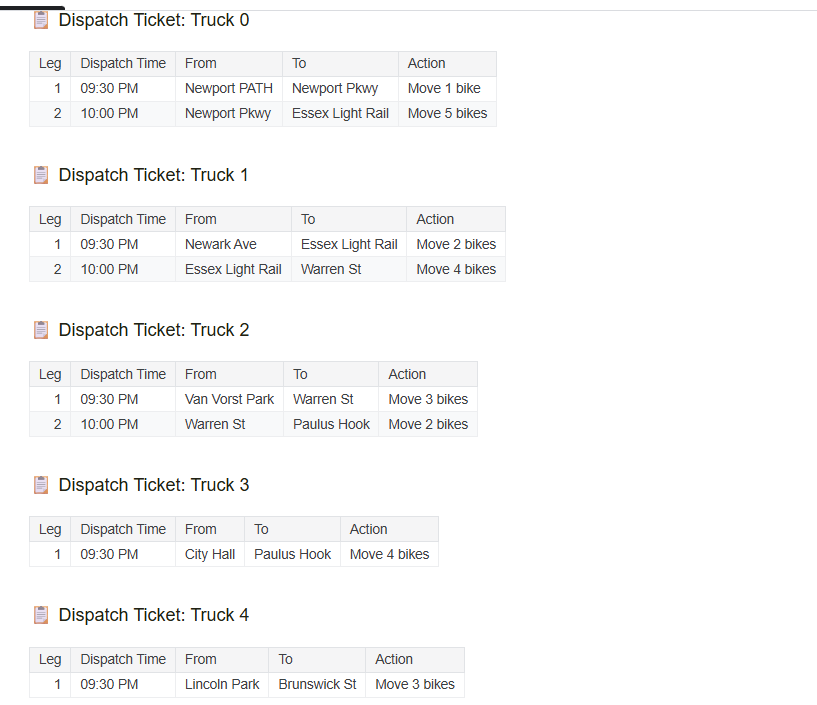}
    \caption{A representative output from the Agentic AI layer, showing the Manager's Briefing and a sample of the detailed, per-truck Dispatch Tickets.}
    \label{fig:smartflow_simulation}
\end{figure}

The final outputs of the framework translate this complex strategy into an actionable and interpretable format. The agentic AI layer successfully generated clear, factually grounded reports for each run, including a high-level "Manager's Briefing" and detailed, per-truck dispatch tickets for operators (Fig.~\ref{fig:agentic_report}). This seamless communication is complemented by a final, interactive map that provides a holistic visualisation of the optimised truck routes and the rebalanced network (Fig.~\ref{fig:smartflow_simulation}).

\begin{figure*}[H]
 \centering
 \includegraphics[width=0.9\textwidth]{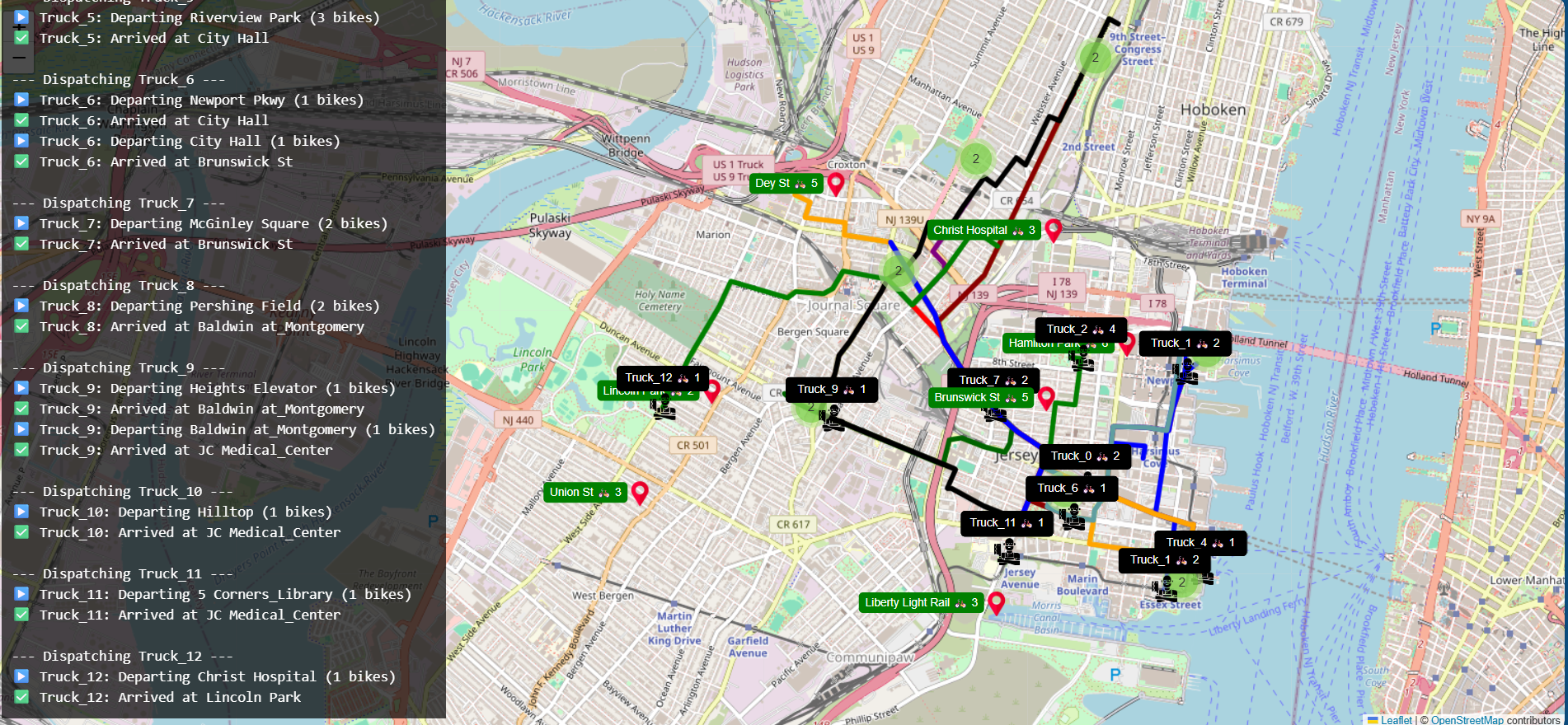}
 \caption{A snapshot of the final interactive map from a representative run, showing the optimised truck routes and the rebalanced final state of the network.}
 \label{fig:final_map}
\end{figure*}


\section{Discussion}
\label{sec:discussion}

The results demonstrate that SmartFlow's hybrid architecture is an effective approach to the dynamic rebalancing problem. The design, however, represents a series of deliberate trade-offs between optimality, scalability, and interpretability, which warrant further discussion.

The decision to use a model-free Reinforcement Learning agent over traditional optimisation methods like Mixed-Integer Linear Programming (MILP) was a primary architectural choice. While MILP can produce provably optimal solutions for static problems~\cite{benchimol2011balancing, chemla2013bike}, it faces exponential growth in computational complexity in dynamic, large-scale environments. RL, by contrast, offers superior scalability and adaptability, learning a robust policy from experience that can generalise to unseen situations without needing an explicit model of the environment's dynamics \cite{sutton2018reinforcement}. The trade-off is moving from guaranteed optimality to a highly effective, learned policy that is more suitable for real-time applications.

Furthermore, SmartFlow decouples high-level strategy from low-level tactics. The RL agent is not burdened with calculating precise vehicle routes; instead, it focuses solely on the strategic question of which stations need servicing. This simplifies the agent's action space and promotes faster learning. The resulting strategic plan is then handed to a deterministic planning module that solves the less complex logistical problem of finding efficient routes. This hybrid design leverages the strengths of both learning-based and algorithmic approaches.

Finally, the integration of the agentic AI layer is a direct response to the "black box" problem often associated with deep learning models. While the DQN agent is highly effective, its internal decision-making process is not inherently transparent. The agentic layer serves as an essential interpretability bridge. By translating the numeric plan into a natural-language report, it makes the system's final output auditable and immediately useful for human operators, a critical factor for building trust and facilitating adoption in real-world logistical operations.

In summary, the SmartFlow architecture was deliberately engineered to balance the exploratory power of modern deep reinforcement learning with the pragmatic requirements of real-world logistical systems. By separating strategy from tactics and grounding communication in verifiable data, the framework offers a solution that is not only effective and scalable but also interpretable and robust. This hybrid design provides a generalisable blueprint for intelligent urban mobility systems where autonomous decision-making must coexist with human operational oversight.


\section{Conclusion and Future Work}

\subsection{Conclusion}
In this paper, we designed, implemented, and rigorously evaluated SmartFlow, a novel framework that successfully integrates deep reinforcement learning with an agentic AI layer to address the Dynamic Bike Rebalancing Problem. Our rigorous evaluation, conducted across multiple independent, seeded runs, confirmed the framework's efficacy. The RL agent learned a robust policy that achieved a remarkable average \textbf{95.47\% reduction in network imbalance} with exceptional operational efficiency, requiring only \textbf{37.21 km of total fleet travel} and achieving a \textbf{Truck Utilisation Rate of 66.15\%}.

The primary contribution of this work is the empirical demonstration of a synergistic pipeline that combines the predictive power of deep RL with the communication capabilities of an agentic AI, successfully bridging the gap between machine intelligence and human execution. By separating high-level strategy from low-level tactics and grounding communication in verifiable data, SmartFlow offers a solution that is not only effective and scalable but also interpretable and robust. This work serves as a strong proof-of-concept for a new generation of intelligent transportation systems where autonomous decision-making must coexist with human operational oversight.

\subsection{Future Work}
The current framework provides a strong foundation for numerous exciting extensions. Future work will focus on bridging the gap between simulation and real-world deployment, and enhancing the model's intelligence and scalability.

\paragraph{Live Data Integration and Real-Time Adaptability.}
A primary goal is to evolve SmartFlow into a real-time operational tool. This would involve integrating live data feeds for station inventory from BSS APIs and incorporating real-time traffic data to generate more accurate travel time estimates. An agent trained in such a dynamic environment would learn a policy that is inherently more robust to unforeseen events.

\paragraph{Advanced Spatio-Temporal Demand Forecasting.}
To improve the agent's proactive capabilities, a dedicated predictive model could be integrated. A deep learning model, such as a Graph Neural Network (GNN), could be used to capture the complex spatial and temporal relationships between stations, providing the RL agent with a richer, forward-looking state representation.

\paragraph{Transition to Multi-Agent Reinforcement Learning (MARL).}
To address the scalability limitations of the current single-agent architecture, we plan to explore a MARL framework. In such a system, each rebalancing truck would be modelled as an independent, cooperative agent learning a decentralised policy. This approach is better suited for city-scale deployments as it distributes the computational load and can handle partial observability.

\paragraph{Enriched Multi-Objective Reward Functions.}
Finally, the agent's reward function could be enhanced to optimise for more complex, multi-objective business goals. Beyond simply balancing the network, the reward signal could be augmented to include penalties for operational costs (e.g., fuel), incentives for environmental sustainability (e.g., prioritising electric vehicles), or bonuses for improving service equity for underserved areas.

\bibliographystyle{plain}
\bibliography{References.bib}
\end{document}